\def\year{2020}\relax
\documentclass[letterpaper]{article} % DO NOT CHANGE THIS
\usepackage{aaai20}  % DO NOT CHANGE THIS
\usepackage{times}  % DO NOT CHANGE THIS
\usepackage{helvet} % DO NOT CHANGE THIS
\usepackage{courier}  % DO NOT CHANGE THIS
\usepackage[hyphens]{url}  % DO NOT CHANGE THIS
\usepackage{graphicx} % DO NOT CHANGE THIS
\usepackage{graphicx}  % DO NOT CHANGE THIS
\title{Convolutional Hierarchical Attention Network for Query-Focused Video Summarization}
\author{
\textbf{Shuwen Xiao\textsuperscript{\rm 1}, Zhou Zhao\textsuperscript{\rm 1}\thanks{Corresponding author.}, Zijian Zhang\textsuperscript{\rm 1}, Xiaohui Yan\textsuperscript{\rm 2}, Min Yang\textsuperscript{\rm 3}}\\ 
\textsuperscript{\rm 1}College of Computer Science and Technology, Zhejiang University, Hangzhou, China\\
\textsuperscript{\rm 2} CBG Intelligent Engineering Dept., Huawei Technologies, China\\
\textsuperscript{\rm 3} Shenzhen Institutes of Advanced Technology (SIAT), Chinese Academy of Sciences\\
\{xiaoshuwen,zhaozhou,ckczzj\}@zju.edu.cn, yanxiaohui2@huawei.com, min.yang@siat.ac.cn\\
}

\begin{document}
\maketitle
\begin{abstract}
Previous approaches for video summarization mainly concentrate on finding the most diverse and representative visual contents as video summary without considering the user’s preference. This paper addresses the task of query-focused video summarization, which takes user’s query and a long video as inputs and aims to generate a query-focused video summary. In this paper, we consider the task as a problem of computing similarity between video shots and query. To this end, we propose a method, named Convolutional Hierarchical Attention Network (CHAN), which consists of two parts: feature encoding network and query-relevance computing module. In the encoding network, we employ a convolutional network with local self-attention mechanism and query-aware global attention mechanism to learns visual information of each shot. The encoded features will be sent to query-relevance computing module to generate query-focused video summary. Extensive experiments on the benchmark dataset demonstrate the competitive performance and show the effectiveness of our approach.
\end{abstract}

\section{Introduction}
Recently, there is an emerging research direction in the task of video summarization, which is query-focused video summarization~\cite{sharghi2017query,zhang2018query,vasudevan2017query,wei2018video}. Given a long video and a user query which is some concepts, the goal of query-focused video summarization is not only to remove the redundant parts of the video, to find key frame / shots in the video, but also to pick out those segments that related to the user's query. It can be helpful in some scenarios such as when a user wants to customize a brief summary from his daily video logs.

\begin{figure}[t]
\centering
\includegraphics[width=0.9\linewidth]{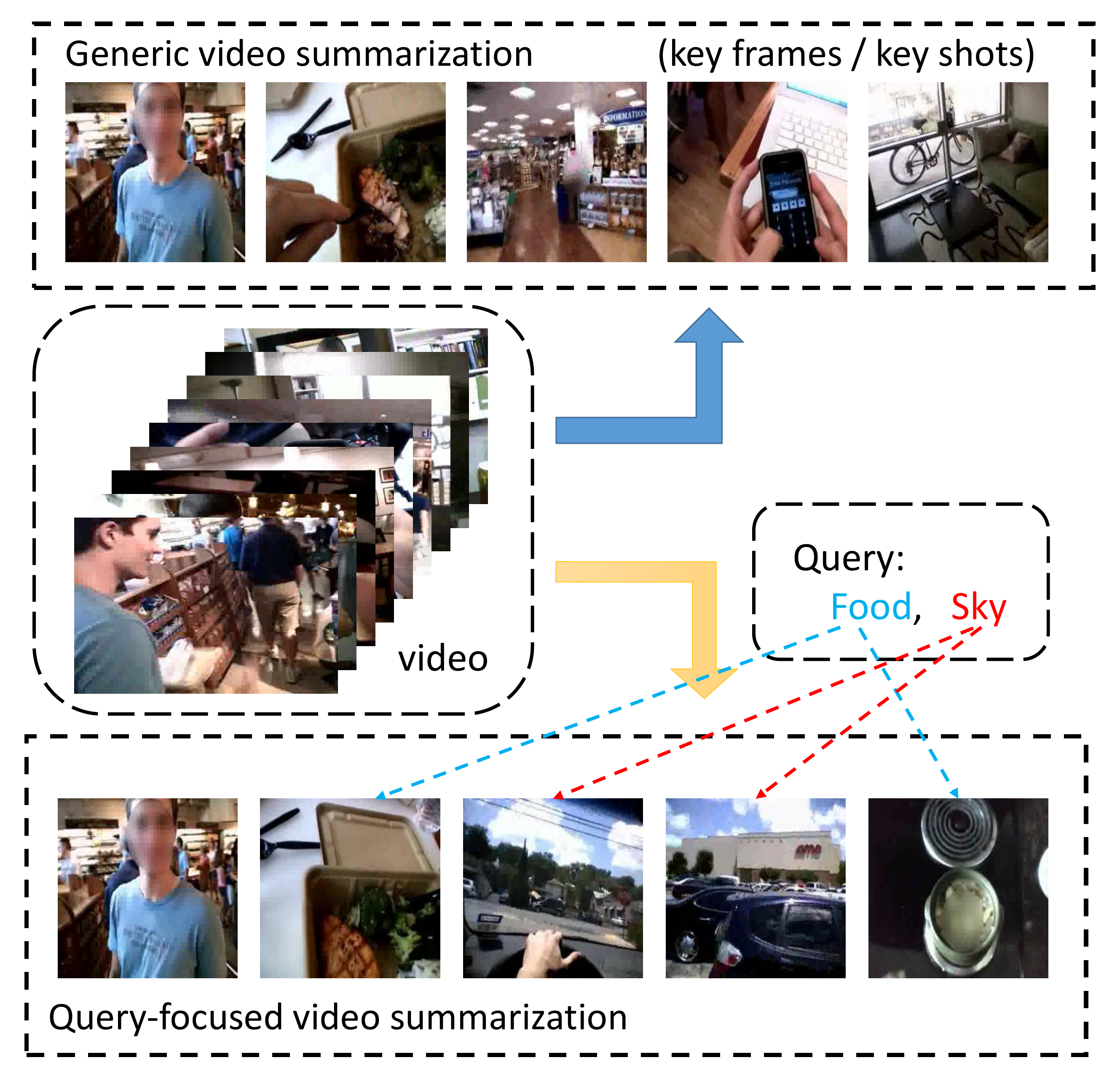}
\caption{The goal of generic video summarization is to select a compact subset of frames or shots that remains the important information from the original video, while the task of query-focused video summarization takes user's query into consideration and generate query-related video summaries.}
\label{fig:demo}
\end{figure}

Generic video summarization is to generate a compact version of the original video by selecting the highlights of the long video and eliminating the trivial content. Automated video summarization techniques are useful in many domains, such as motion recognition, surveillance video analysis, visual diary creation for personal lifelog videos, and video previews of video sites. Besides, modules for video summarization can also be implemented as an intermediate component to assist downstream tasks. As is shown in figure~\ref{fig:demo}, there are three differences between query-focused video summarization and generic video summarization. Firstly, the video summary needs to take  the subjectivity of users into account, as different user queries may receive different video summaries. Secondly, trained video summarizers cannot meet all the users' preferences and the performance evaluation is often to measure the temporal overlap, makes it hard to capture the semantic similarity between summaries and original videos~\cite{sharghi2017query}. Thirdly, the textual query will bring additional semantic information to the task. 

Recent researches on query-focused video summarization~\cite{sharghi2017query,zhang2018query} mainly consider the task as sequence labeling problem and they employ models based on sequential structure such as long short-term memory (LSTM)~\cite{hochreiter1997long}. At each time step, the model outputs a binary score to predict whether this frame is important and related to the given query. However, sequential models usually perform the computation step by step and when the length of video increases, the computing time for the models also become longer. Moreover, sequential models cannot easily handle long distance relation among the video when it is too long, because of the problem of vanishing gradients. In ~\cite{rochan2018video}, they use fully convolutional networks as an encoder-decoder structure to solve video summarization task, which shows the ability of convolutional networks to generate high-quality video summaries.

In this paper, we consider the task of query-focused video summarization as a ranking problem, where we first select out the important visual content and then compute the similarity between visual content and the given query. We propose a novel method, named Convolutional Hierarchical Attention Network (CHAN), which consists of two parts: 1) a feature encoding network to learn features from each video shot in parallel from a local perspective and a global perspective; 2) query-relevance ranking module to calculate the similarity score with respect to a query for each shot and then select video content related to the given query. In feature encoding network, we utilize fully convolutional structure to decrease the dimension on the temporal sequence and the dimension of visual feature, in order to reduce the number of parameters in the model. To get a better comprehension of video, we first split video into segments and employ an information fusion layer in the encoding network, which learns visual representation inside a segment and between several segments. Combining local-level self-attention and query-aware global-level attention mechanism, information fusion layer produce features with different level information. Then, we propose a query-relevance ranking module, which can handle the modality between video and text query and compute the similarity score for each shot. Finally, the model generates query-related video summary with the similarity score.

The main contributions of this paper are summarized as follows:

\begin{itemize}
\item We propose a novel method named Convolutional Hierarchical Attention Network (CHAN), which is based on convolution network and global-local attention mechanism and able to generate video query-related summary in parallel.

\item We present a feature encoding network to learn the features of each video shot. Inside the feature encoding network, we employ fully convolutional network with local self-attention mechanism and query-aware global attention mechanism to obtain features with more semantic information.

\item We employ a query-relevance computing module which takes the feature of video shot and query as input and then calculate the similarity score. Query-related video summary is generated based on similarity score.

\item We perform extensive experiments on the benchmark datasets to demonstrate the effectiveness and efficiency of our model.
\end{itemize}

The rest of this paper is organized as follows. In Section~\ref{sec:related}, we provide a review of the related work about generic video summarization and query-focused video summarization. In the following, Section~\ref{sec:model} introduces each component of our approach. A variety of experimental results are presented in Section~\ref{sec:exp}. Finally, we provide some concluding remarks in Section~\ref{sec:conclusion}.

\section{Related Works} \label{sec:related}
The goal of query-focused video summarization is obtaining a diverse subset of video frames or video segments that are not only related to the given query but also contains the original information of video. It can be categorized into two domains: (1) generic video summarization (2) query-focused video summarization.

\subsection{Generic Video Summarization}
Generic video summarization has been studied for many years and several methods have been proposed so far. They can be roughly categorized into three aspects: unsupervised, supervised or weakly supervised approaches.

Unsupervised video summarization approaches\cite{ngo2003automatic,hong2009event,khosla2013large,zhao2014quasi,panda2017weakly} take advantage of specific selection criteria to measure the importance or interestingness of video frames and then generate video summaries. These methods have conventional methods, that is, methods that use some low-level video features for calculation, as well as more recent methods, which combine deep learning. Conventional unsupervised video summarization approaches mostly use hand-crafted heuristics~\cite{ngo2003automatic,zhao2014quasi} (such as diversity and representativeness) or frame clustering method~\cite{otani2016video,wang2016video} to decide whether to choose a video frame as a keyframe. Assuming that Web images bring extra visual information which contains people's interest, some methods~\cite{chu2015video,khosla2013large,panda2017weakly} use the web images to generate user-oriented video summaries. In~\cite{kanehira2017viewpoint}, Kanehira \emph{et al.} take $viewpoint$ into consideration and calculate the video-level similarity between multiple videos to generate video summary. More recently, some methods based on deep learning have been proposed. In ~\cite{mahasseni2017unsupervised}, Mahasseni \emph{et al.} propose a generative adversarial network. The model consists of summarizer and discriminator. The summarizer selects a subset of keyframes and then reconstruct the video and the discriminator is to minimize the distance between the information of reconstructed videos and original videos. Zhou \emph{et al.}~\cite{Zhou2017Deep} propose a reinforcement learning algorithm to summarize videos, by designing the diversity-representativeness reward and applying algorithm based on policy gradient to optimize the model. 

Supervised video summarization methods~\cite{gygli2014creating,potapov2014category,song2015tvsum,gygli2015video,zhang2016video,rochan2018video} have emerged in recent years, and the training data used in these methods contains raw videos and human-created ground-truth annotations. Combined with a large number of human annotations, models can capture the video content with more semantic information, so these approaches are usually superior to unsupervised methods. In ~\cite{gygli2014creating}, video summarization is formulated as an interestingness scoring problem and frames with higher scores are selected as summaries. Some approaches utilize the additional information from web images ~\cite{khosla2013large}, categories~\cite{potapov2014category}, and titles\cite{song2015tvsum} to improve the quality of video summaries. In ~\cite{gygli2015video}, Gygli \emph{et al.} propose an adapted submodular function with multiple objects to tackle the problem of video summarization. Yao \emph{et al.}\cite{Yao_2016_CVPR} employ a deep rank model, taking a pair of highlight segment and non-highlight segment as inputs, to rank each segment. In ~\cite{zhang2016video}, Zhang \emph{et al.} implement the dppLSTM model by combining determinantal point process and bi-directional LSTM to predict whether a frame is a keyframe. In ~\cite{zhao2018hsa}, the author proposed a hierarchical structure-adaptive RNN which is composed of two layer LSTM, where the first layer is aimed to split the video into several shots then the second layer learns shot-level information and obtain shot-level score. In ~\cite{rochan2018video}, Rochan \emph{et al.} propose a fully convolutional sequence network, of which the structure is based on convolutional layer and deconvolutional layer, consequently the network runs in parallel.

Although the supervised video summarization method performs better than the unsupervised method, there still exist some weaknesses. Firstly, getting ground-truth annotations in a video summarization dataset is time-consuming and labor-intensive. Secondly, in reality, the results returned by the video summary generator should not be unique, but the models trained with supervised methods can lead to overfitting and thus are difficult to generalize. To overcome these weaknesses, some weakly supervised methods have been proposed. These methods typically utilize readily available label, such as video categories, as additional information to improve model performance. In ~\cite{panda2017weakly}, a 3D ConvNet is trained to predict the category of the video. The importance score of each frame is obtained by calculating the back-propagated gradient returned by the true category. In ~\cite{cai2018weakly}, an encoder-attention-decoder structure is built, where the encoder learns the latent semantics from web videos and the decoder generate the summary.

\subsection{Query-Focused Video Summarization}
Compared to the generic video summarization, the task of query-focused video summarization takes user's query into account. Specifically, the dataset provides concept annotations for each shot in the video and contain more semantic information than the general form. 

Some recent works take semantics into consideration when tackling the task of video summarization. In ~\cite{wei2018video}, they create video descriptions for videos in summarization datasets and train model based on the descriptions to leverage the additional textual information. In ~\cite{sharghi2016query}, Sharghi \emph{et al.} introduce a DPP-based algorithm, which takes user query as input and generates query-focused video summary. In ~\cite{sharghi2017query} explore query-focused video summarization, which generates a summary based on video content and user query and proposes a memory network based model. Vasudevan \emph{et al.}~\cite{vasudevan2017query} introduce a quality-aware relevance model with submodular maximization to select important frames. In ~\cite{zhang2018query}, the author proposes generative adversarial network to tackle this challenge, where the model will generate a query-focused summary, a random summary. With the ground truth summary, a three-player loss is introduced to optimize the model. These approaches are sequential model, using DPP-based algorithm or LSTM structure. To the best our knowledge, we are the first to propose an approach based on self-attention mechanism in query-focused video summarization tasks. Our proposed model is built based on convolutional neural network and attention mechanism, which can perform computation in parallel. With the query-relevance computing module, our model can produce query-related video summary.

\begin{figure*}[t]
\centering
\includegraphics[width=0.9\textwidth]{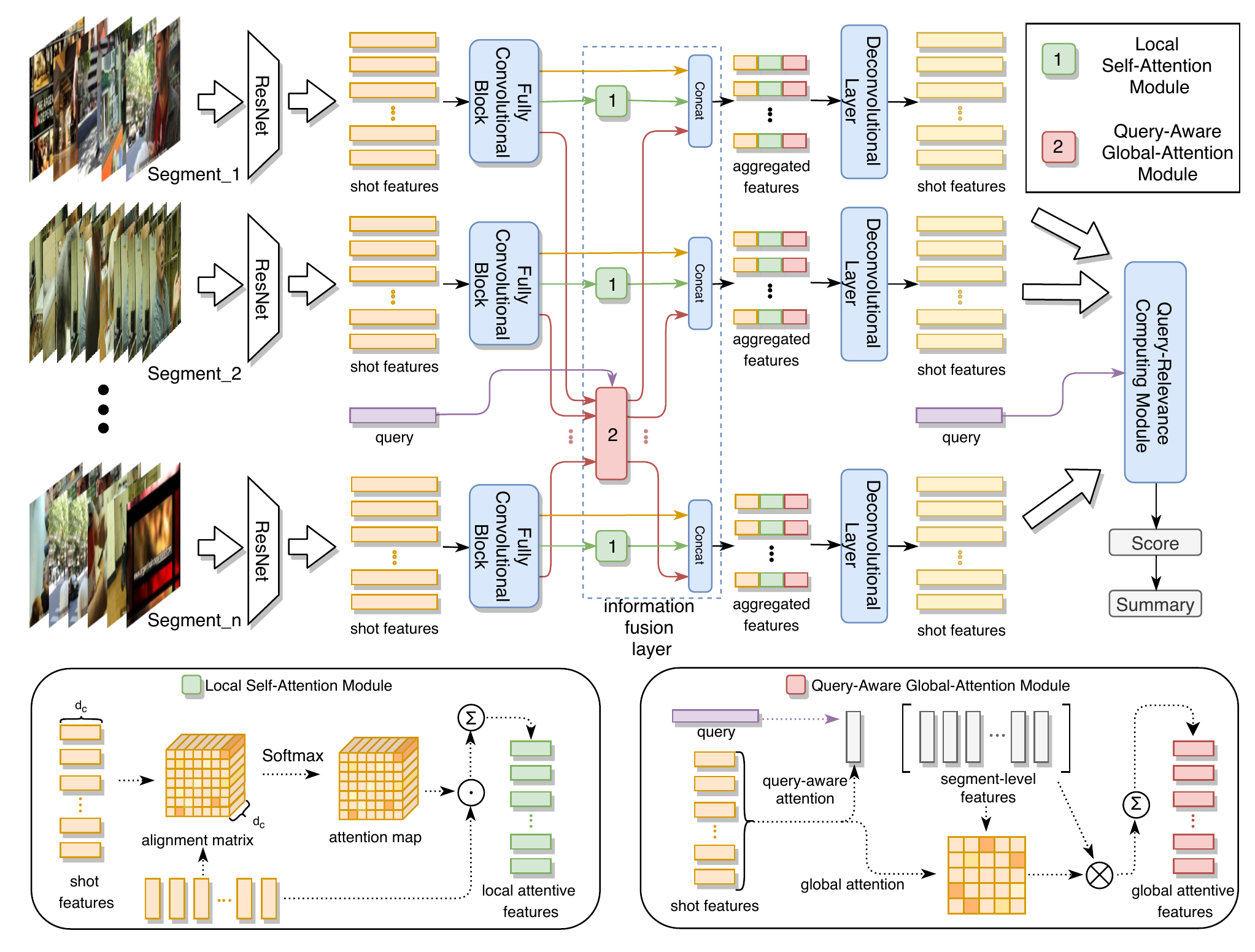}
\caption{The framework of our proposed convolutional hierarchical attention network(CHAN). We first split the original long video into several segments and then extract visual features using pretrained network. Then we send the visual features to feature encoding network to learning shot-level features. In the feature encoding network, we propose a local self-attention module to learn the high-level semantic information in a video segment and employ a query-aware global-attention module to handle the semantic relationship between all segments and the given query. In order to reduce the parameter in the model, we employ fully convolutional block to decrease the dimension of visual feature and shorten the length of the shot sequence. Finally, all the aggregated shot-level visual features will be sent to query-relevance ranking module to obtain similarity score and the query-related video summary will be generated based on the similarity score.}
\label{fig:framework}
\end{figure*}

\section{Method}~\label{sec:model}
In this section, we describe the task of query-focused video summarization and illustrate the details of our proposed model.

\subsection{Problem Formalization}
Given a long video {\bf v} and a query {\bf q}, the goal of query-focused video summarization is to find out the diverse and representative subset of query-related frames or video shots. To tackle with this challenge, we denote the task as a problem of calculating the shot-query similarity. First, we denote the video as a sequence of video shots $(s_1, s_2, \dots, s_n)$, where $n$ is the number of video shots. Each video shot is a small clip of the original video. Then we denote the representation of query as $h_q$. Moreover, in the benchmark dataset ~\cite{sharghi2017query}, in where each query is consist of two concept $(c_1, c_2)$, we have to compute concept-related score for each shot and then merge two kind of score as the query-related score. Finally, based on the score, we can produce a diverse subset of video segments which can not only represent the origin video but related to the query.

\textbf{Input:} A long video ${\bf v}$ and a query ${\bf q}$.

\textbf{Output:} A diverse subset of video shots that remains the origin information of original video and each shot in the subset is related to the given query.

\subsection{Feature Encoding Network}
In this part, we will introduce each components of feature encoding network, as is shown in figure \ref{fig:framework}. The main idea of our approach is that we build a encoding model which can not only capture the relation between each two shots in local-level and global-level to find out important shot in parallel, but also search for shots that related to the query. We then propose the query-relevance computing module to compute the similarity score between encoded shot feature and query. We use convolutional structure and attention mechanism, in which the computational processes are parallel and can capture multi-distance visual information.

Considering the hierarchy of video, a long video consists of several stories, and a story contains multiple segments, where each segment is composed of many frames. In order to get better comprehension of long videos, we need to learn the representation of video from different levels, which means we need to split the video into small video segments. There are some heuristic algorithms like uniform segmentation and the method based on visual feature~\cite{chu2015video}. We construct video segments using the Kernel Temporal Segmentation~\cite{potapov2014category}, which preserves visual consistency in each segment.

\subsubsection{Fully Convolutional Block}
After the segmentation process, we then use the pretrained ResNet to extract the visual feature of each shot. We denote the shot-level features in a specific video segment as $({\bf v}_{1},{\bf v}_{2},\ldots,{\bf v}_{s_k})$, where $s_k$ is the number of shots in the video segment. We propose a convolutional structure to further encode the visual feature.

To build a model with better vision of video shots in local area and parallel computation process, we use 1D fully-convolutional network architecture~\cite{long2015fully} to encode the shot-level visual feature. Moreover, for the purpose of handling long distance among the video segment, we utilize dilated convolutions to obtain larger receptive field. We first propose convolutional networks with different filter size and then concatenate their outputs which enables the model to receive more information. The dilated convolution operation on $i$-th shot in a video segment is defined as,
\begin{equation}
    o_i = \sum_{t=-k}^k{f(t) \cdot {\bf v}_{i+d \cdot t}}
\label{Convolution}
\end{equation}
where $2k+1$ is the filter size, $f$ is the filter and $d$ is dilation factor. With different filter sizes and large dilation, we can obtain outputs that represent more information from inputs. Then we employ a pooling layer on the temporal axis of the video, which can reduce the computing time and also decrease the running memory of the model. We connect different fully convolutional block and construct a multi-layer block to extracted representative features.

\subsubsection{Information Fusion Layer}
We denote the output features of fully convolutional block as $(\hat{v}_1, \hat{v}_2, \ldots, \hat{v}_t)$, where $t$ is length of output feature sequence, and the output features are the inputs of the information fusion layer. The outputs of information fusion layer are a sequence of concatenated vectors of outputs from previous block, local attentive representations and query-aware global attentive representations.

The local self-attention module is designed to capture the semantic relations between all shots among a video segment. The detailed structure is presented in the left bottom of figure~\ref{fig:framework}. Given the sequence of output features $(\hat{v}_1, \hat{v}_2, \ldots, \hat{v}_t)$ from fully convolutional block, we can compute the alignment matrix. The alignment score vector for $i$-th element and $j$-th element is given by
\begin{equation}
    f(\hat{v}_i, \hat{v}_j) = {\bf P} {\rm tanh} ({\bf W}_{1} \hat{v}_i + {\bf W}_{2}\hat{v}_j + {\bf b}) \label{local self attention matrix},
\end{equation}
where ${\bf P}, {\bf W}_{1} , {\bf W}_{2} \in {R}^{{d}_{c} \times {d}_{c}}$ are the trainable parameters, ${\bf b} \in {R}^{{d}_{c} \time 1}$ is bias vector and $d_c$ is the dimension of feature vector $\hat{v}_i$. The score vector $f(\hat{v}_i, \hat{v}_j) \in {R}^{d_c}$ and the shape of output alignment score matrix is $t \times t \times {d}_{c}$, which means the module can focus on the representative features from a temporal perspective and  from the viewpoint of each dimension of the features. We propose this module to learn the relative semantic relationship of different frames in the same segments, and for different segments, the relation structure should be similar. Therefore, for different video segments, our local self-attention modules share all the trainable parameters, which also reduces the amount of parameters in our model. The local attentive video feature for $i$-th element is computed by,
\begin{eqnarray}
    \gamma_{ij} & = & \frac{\exp(f(\hat{v}_i, \hat{v}_j))} {\sum_{k=0}^t \exp(f(\hat{v}_i, \hat{v}_k))} \\
    {\hat v}_i^l & = & \sum_{j=0}^t \gamma_{ij} \odot \hat{v}_j \label{local self attention},
\end{eqnarray}

Then we introduce the query-aware global attention module which is to model the relationship of different video segments among the video and to generate query-focused visual representation, as is shown in the right bottom of figure~\ref{fig:framework}. First, given a sequence of visual representation generated from the fully convolutional block $(\hat{v}_1, \hat{v}_2, \ldots, \hat{v}_t)$ and a query $q$, which is composed of two concept $(c1, c2)$, we can calculate the query-aware segment-level visual feature for this video segment by,
\begin{eqnarray}
e_{i} & = & v^T {\rm tanh} ({\bf W}_{1} \hat{v}_i + {\bf W}_{2}{\bf h}^q + b)  \\
\gamma_{i} & = & \frac{\exp(e_{i})} {\sum_{k=0}^t \exp(e_{k})} \\
v^{(s)} & = & \sum_{i=0}^t \gamma_{i} \hat{v}_i
\end{eqnarray}

where $v^T, {\bf W}_{1} , {\bf W}_{2}$ are the trainable parameters, $b$ is bias vector and $h^q$ is the average of representation of concepts. The query-aware segment-level features represent aggregated representation for a video segment to a specific query. Then we compute the query-aware global-attentive representation for each shot. Given the visual feature $\hat{v}_i$ and all the segment-level visual representation $(v_1^{(s)}, v_2^{(s)}, \ldots, v_{m}^{(s)})$, where $m$ is the number of video segments, the global-attentive representation for $i$-th shot can be computed by,
\begin{eqnarray}
e_{j}^g & = & v^T {\rm tanh} ({\bf W}_{1}^{g} \hat{v}_i + {\bf W}_{2}^{g} v_{j}^{(s)} + b)  \\
\gamma_{j}^g & = & \frac{\exp(e_{j}^g)} {\sum_{k=0}^m \exp(e_{k}^g)} \\
{\hat v}_i^g & = & \sum_{j=0}^m \gamma_{j}^g v_{j}^{(s)}
\end{eqnarray}

Finally, we concatenate three kinds of features as the outputs of the information fusion layer. More specifically, the fused feature for the $i$-th item is $\hat{v}_i^f = [\hat{v}_i; \hat{v}_i^l; \hat{v}_i^g]$.

\subsubsection{Deconvolutional Layer}
We use fully convolutional block to decrease the number on temporal axis of video features and we propose several 1D deconvolutional layers to recover the original number of video shots.

\subsection{Query-Relevance Computing Module}
In this part, we are going to introduce our query-relevance computing module, which can compute the similarity score between video shot and user's query.

Due to the fact that a query is composed of two concepts, our goal is to learn a module that can select the most concept-related shots by returning a concept-relevant score. The module takes the shot-level features generated by feature encoding network, which are denoted as $(\hat{v}_1^f, \hat{v}_2^f, \ldots, \hat{v}_n^f)$ and concepts as inputs. Moreover, each segment is related to one or multiple concepts. Given a specific concept $c$, we first obtain its embedding feature $f_c$ using pretrained language model. Given a concept feature $f_c$ and shot-level features $\hat{v}_i^f$ of $i$-th shot, we first calculate their distance-based similarity by

\begin{eqnarray}
d_i = {\bf W}_{f}\hat{v}_i^f \odot {\bf W}_{c}{f}_{c}
\end{eqnarray}

where ${\bf W}_f$ and ${\bf W}_c$ are the parameter matrices which project the visual features and textual features into the same vector space. Then we let the output pass a MLP and get the concept-relevant score between $i$-th segment and concept $c$. The average of two concept-relevant score is taken as the query-relevant score $s = \{s_1, s_2, \dots, s_n\}$.

Given the ground truth annotations ${\hat s} = \{{\hat s}_1, {\hat s}_2, \dots, {\hat s}_n\} \in [0, 1]$ which represents whether a shot is related to a concept, the loss can be defined as:
\begin{eqnarray}
L_{summ} = \frac{1}{T}\sum_{t=1}^{T}{{\hat s}_{t}log{s}_{t} + (1-{\hat s}_{t})log(1-{s}_{t})}
\end{eqnarray}

By minimizing the loss, query-relevance computing module can focus on the most concept-related video shots.

\section{Experiments}~\label{sec:exp}
\subsection{Datasets and Experiment Settings}
\subsubsection{Datasets}
We evaluate our method on the query-focused video summarization dataset proposed in ~\cite{sharghi2017query}. The dataset is built upon the UT Egocentric(UTE) dataset~\cite{lee2012discovering}, which contains videos taken from the first person perspective. The dataset has 4 videos containing different daily life scenarios, each of which lasts 3$\sim$5 hours. In ~\cite{sharghi2017query}, they provide a set of concepts for user's queries, in which the total number of concepts is 48. The concepts are concise and diverse, which are related to some common objects in our daily life. Each query is composed of two concepts and there are 46 queries in the dataset. As for queries, there are four different scenarios in this task~\cite{sharghi2017query}, that is, 1) all concepts in the query appear in the same video shot, 2) all concepts in the query appear in the video but not in the same shot, 3) some of the concepts in the query appear in the video, 4) none of the concepts in the query appear in the video. It is worth mentioning that the fourth scenario is to some extend the same as general form video summarization. The dataset provides per-shot annotation, from which each shot is labeled with several concepts.

\subsubsection{Data Preprocessing}
We preprocess the videos in query-focused video summarization dataset as follow. To get the shot features, we sample the video to 1 fps, then reshape the size of all frames to $224 \times 224$. Next, we extract visual representation of each frame with the pretrained ResNet~\cite{szegedy2017inception} which is pretrained on ImageNet~\cite{deng2009imagenet}, and taking the 2048-dimensional vector for each frame. Similar to the setting in ~\cite{sharghi2017query}, we take 5 seconds as a shot and compute the average of each frame from a shot as its shot-level feature. Utilizing the KTS algorithm~\cite{potapov2014category}, we can split video into small segments under the conditions that the number of segments in a video is no more than 20 and the number of shots in a segments is no more than 200. For the concept words in each query, we extract its word representation using
publicly available Glove vectors~\cite{pennington2014glove}.

\subsubsection{Evaluation Protocol}
For fair comparison, we follow the protocols setting in ~\cite{sharghi2017query}. Considering that the video summary should be a relatively subjective result rather than a unique result, the author provides a dataset with dense per-shot annotations. They measure the performance of the model by calculating semantic similarity between different shots, rather than just measuring temporal overlap or relying on low-level visual features. In ~\cite{sharghi2017query}, they first calculate the conceptual similarity between each two video shots based on the intersection-over-union(IOU) of their related concepts. Then they use the conceptual similarity as the edge weights in the bipartite graph between two summaries to find the maximum weight matching of the graph. Finally, precision, recall and F1 score can be computed based on the number of matching pairs.

\subsection{Implementation Details}
We use Pytorch to implement our approach on a server with a GTX TITAN X card. In the feature encoding layer, we propose two-layer fully convolution block, in which the output channel dimension for first layer is 256 and for the second one is 512. In the local self-attention module and query-aware global attention module, the dimension of attention $d_c$ is set to 256. The dimension of the visual-textual fused space in query-relevance computing module is 512. Following the setting in ~\cite{sharghi2017query}, we randomly select two videos for training, one for testing and the remaining one for testing. In the training process, we use Adam optimizer~\cite{kingma2014adam} to minimize the loss, with its initial learning rate 0.0001 and decay rate of 0.8. The mini-batch strategy is also used and the batch size is set to 5. After obtaining the similarity score, we create query-related video summary by selecting video shots with the highest score.

\begin{table*}[ht]
\caption{Comparison results on the query-focused video summarization dataset in terms of Precision, Recall and F1-score.}
  \centering
  \resizebox{.95\linewidth}{!}{
  \begin{tabular}{c|ccc|ccc|ccc|ccc|ccc}
    \hline
      &
    \multicolumn{3}{c|}{SeqDPP} &
    \multicolumn{3}{c|}{SH-DPP}  &  
    \multicolumn{3}{c|}{QC-DPP}  & 
    \multicolumn{3}{c|}{TPAN} & 
    \multicolumn{3}{c}{CHAN}\\ 
    \hline
      & Pre &Rec &F1 &  Pre &Rec &F1  &  Pre &Rec &F1  &  Pre &Rec &F1&  Pre &Rec &F1\\
    \hline
    Vid1 &53.43&29.81 &36.59 &50.56&29.64 &35.67 &49.86&53.38 &  48.68 & 49.66 & 50.91 & 48.74 & 54.73 & 46.57 & {\bf 49.14} \\
    Vid2 &44.05&46.65 &43.67 &42.13&46.81 &42.72 &33.71&62.09 &41.66 & 43.02 & 48.73 & 45.30 & 45.92 & 50.26 & {\bf 46.53} \\
    Vid3 &49.25&17.44 &25.26 &51.92&29.24 &36.51 &55.16&62.40 & 56.47 & 58.73 & 56.49 & 56.51 & 59.75 & 64.53 &  {\bf 58.65} \\
    Vid4 &11.14&63.49 &18.15 &11.51&62.88 &18.62 &21.39 & 63.12 &29.96 & 36.70 & 35.96 & {\bf 33.64} & 25.23 & 51.16 & 33.42 \\
    \hline
    Avg. &39.47&39.35 &30.92 &39.03&42.14 &33.38 & 40.03 & 60.25 &44.19 & 47.03 & 48.02 & 46.05 & 46.40 & 53.13 & {\bf 46.94} \\
    \hline
  \end{tabular}
  }
   \label{tab: comparison}
\end{table*}

\subsection{Compared Models}
In our evaluation, we compare our proposed method with other query-focused video summarization approaches as follows: 

\begin{itemize}

\item {\bf SeqDPP} method~\cite{gong2014diverse} which formulates video summarization as a subset selection problem and use submodular maximization to found a good summary. The original approach does not consider user queries. 

\item {\bf SH-DPP}~\cite{sharghi2016query} is the extension of SeqDPP method, where the author add a extra layer in the process of SeqDPP to judge whether a video shot is related to a given query. 

\item {\bf QC-DPP}~\cite{sharghi2017query} is the extension of SeqDPP method. In this approach, the author introduces memory network to parameterize the kernel matrix. 

\item {\bf TPAN}~\cite{zhang2018query} is three-player adversarial network. The author uses generative adversarial network to tackle with the task and introduce a random summary as an extra adversarial sample.
\end{itemize}

\begin{table}[h]
\caption{Ablation analysis on query-conditioned video summarization in terms of Precision , Recall and F1-score.}
\centering
\begin{tabular}{l||c|c|c}
\hline
Model  & Pre & Rec & F1  \\
\hline
CHAN w/o Local Att   & 42.72 & 49.04 & 43.26        \\
CHAN w/o Global Att  & 37.62 & 43.17 & 38.09              \\
\hline
CHAN   & 46.40  & 53.13  & 46.94  \\
\hline
\end{tabular}
\label{tab:ablation}
\end{table}

\subsection{Experimental Results}
Table ~\ref{tab: comparison} show the comparison results for query-focused video summarization in terms of precision, recall and f1-score. We compare our method with other approaches that have been used in this task. It is shown that our method outperforms the state-of-the-art approach (TPAN) by 1.9\%. More specifically, for the video 2 and video 3, we obtain a better performance than TPAN~\cite{zhang2018query}, by 2.64\% and 3.6\% respectively. The improvements of performance identify the effectiveness of our approaches to learn the relevance between the video shots and user's query. The average running time for inference phase of each video is 134.4ms, which is shorter than 1.614s ~\cite{zhang2018query} by 91.6\%, showing the efficiency of our approach for its parallel computing ability.

The ablation study of each components in our model, which is shown in Table ~\ref{tab:ablation}. The F1-score of the model without local self-attention module is reduced by 7.84\%. Moreover, without query-aware global attention module, the performance decreases by 18.8\%. The results illustrate the effectiveness of these two attention modules. Local self-attention module and query-aware global attention module can capture visual information inside a video segment and between segments, which is helpful to improve performance.

\subsection{Qualitative Results}
\begin{figure}[t]
\centering
\includegraphics[width=0.8\linewidth]{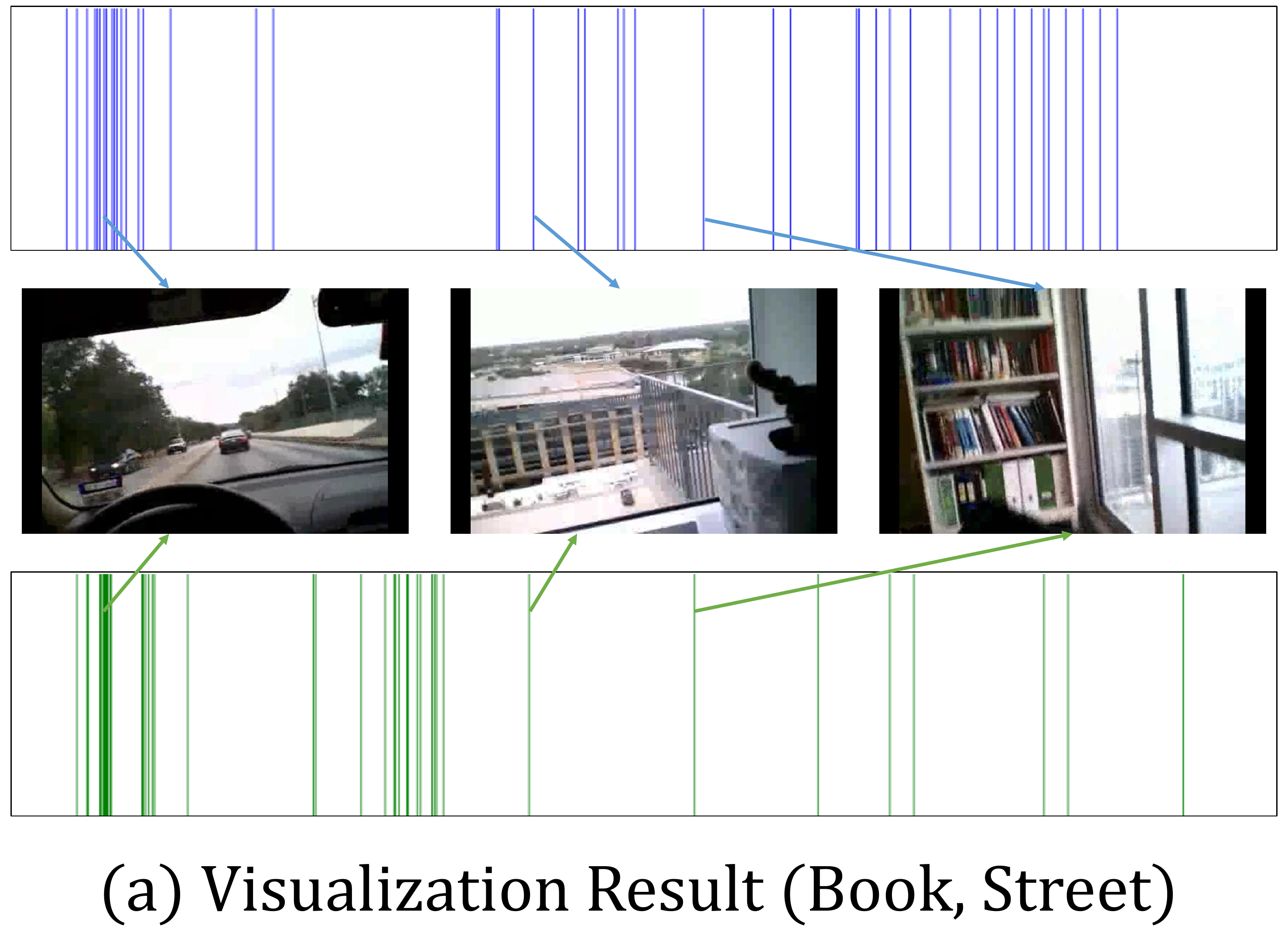}
\caption{The visualization results of our approach. The x-axis denotes the video shot number. Green lines represent the ground truth annotations and blue lines represent the predicted key shots from our method. Figures shows the summary for the query ``Book Street''}
\label{fig:example}
\end{figure}

We present the visualization result in Figure ~\ref{fig:example}. We take one user queries, ``Book Street'', as the inputs of our model. The x-axis in the figure represents the temporal order of video shots. The green lines represent the ground truth annotations of key shots which are related to the query. The blue lines denote the results of predicted key shots. It can be observed that the selected summaries are related to one or both concepts in the given query, which demonstrates that our proposed method is able to find diverse, representative and query-related summaries.

\section{Conclusion} \label{sec:conclusion}
In this paper, we study the task of query-focused video summarization. We formulate query-focused video summarization as a task of computing relevance score between video shot and query, then we generate video summary based on the relevance score. We propose a Convolutional Hierarchical Attention Network which firstly encodes the video shots in parallel, then computes relevance scores between each shot and query and finally generate query-related video summary. Extensive experiments on the benchmark dataset show the effectiveness and efficiency of our approach. Moving forward, we are going to design a more general model which can generate video summary for new queries.

\section{Acknowledgments}
This work is supported in part by the National Natural Science Foundation of China under Grants No.61602405, No.U1611461, No.61751209, No.61836002 and China Knowledge Centre for Engineering Sciences, and Technology.

\bibliographystyle{aaai}
\bibliography{reference}

\end{document}